\documentclass{article}

\PassOptionsToPackage{numbers, compress}{natbib}
\usepackage[final]{nips_2016}

\usepackage[utf8]{inputenc} 
\usepackage[T1]{fontenc}    
\usepackage{hyperref}       
\usepackage{url}            
\usepackage{booktabs}       
\usepackage{amsfonts}       
\usepackage{nicefrac}       
\usepackage{microtype}      

\usepackage{amsfonts}
\usepackage{amsmath}
\usepackage{amsthm}
\usepackage{color}
\usepackage{subcaption}
\usepackage{graphicx}
\usepackage[ruled,lined,boxed,commentsnumbered]{algorithm2e}
\usepackage{algpseudocode}

\DeclareMathOperator*{\argmin}{arg\,min}

\newcommand{\diag}{\mathrm{diag}}
\newcommand{\Tr}{\mathrm{Tr}}

\newcommand{\bigO}{\mathcal{O}}

\newcommand{\xvect}{\mathbf{x}}

\newcommand{\wvect}{\mathbf{w}}

\title{Mini-Batch Spectral Clustering}

\author{
  Yufei Han \\
  Symantec Research Labs\\
  \texttt{yfhan.hust@gmail.com} \\
  \And
  Maurizio Filippone \\
  EURECOM \\
  \texttt{maurizio.filippone@eurecom.fr} \\
}

\begin{document}

\maketitle

\begin{abstract}

The cost of computing the spectrum of Laplacian matrices hinders the application of spectral clustering to large data sets.
While approximations recover computational tractability, they can potentially affect clustering performance.
This paper proposes a practical approach to learn spectral clustering based on adaptive stochastic gradient optimization.
Crucially, the proposed approach recovers the exact spectrum of Laplacian matrices in the limit of the iterations, and the cost of each iteration is linear in the number of samples.
Extensive experimental validation on data sets with up to half a million samples demonstrate its scalability and its ability to outperform state-of-the-art approximate methods to learn spectral clustering for a given computational budget.

\end{abstract}

\section{Introduction}

Over the past two decades, spectral clustering has established itself as one of the most prominent clustering methods \citep{Shi00,Ng02}.
The effectiveness of spectral clustering in identifying complex structures in data is a direct consequence of its close connection with kernel machines \citep{Dhillon04}.
Because of this connection, however, it is also apparent that spectral clustering inherits the scalability issues of kernel machines.
In spectral clustering, the computational challenge is to determine the spectral properties of the so called Laplacian matrix \citep{vonLuxburg07}.
Denoting by $n$ the number of samples, storing the Laplacian matrix requires $\bigO(n^2)$ space while calculating the spectrum of the Laplacian requires $\bigO(n^3)$ computations. 

Several approaches have been proposed to reduce the complexity of spectral clustering, such as employing power methods to identify the principal eigenvectors of the Laplacian \citep{Boutsidis15}.
While this approach is exact in the limit of iterations and does not require storing the Laplacian, the complexity is dominated by the iterative multiplication of the Laplacian matrix by vectors, leading to $\bigO(n^2)$ computations.
In order to further reduce this complexity to $\bigO(n)$, a number of approximations are proposed in the literature. 
A popular technique based on the Nystr\"om approximation relies on a small set of inducing points to approximate the spectrum of the Laplacian matrix \citep{Fowlkes04}. 
Other recent approximations which attempt to compress the dataset appear in \citet{Yan09} and \citet{Li16}.
These approximations recover tractability and make it possible to apply spectral clustering to large data sets.
However, approximations can generally affect the quality of the clustering solution, as we illustrate in the experiments.

This paper proposes a novel iterative way to solve spectral clustering in $\bigO(n)$, while retaining exactness in the calculation of the spectrum of the Laplacian in the limit of the iterations.
Denoting by $L$ the Laplacian matrix, the idea hinges on the possibility to cast the algebraic problem of identifying its principal eigenvectors as the following trace optimization problem 
\begin{equation} \label{eq:constrained:optimization}
\argmin_{W \in \mathcal{R}^{n \times k}} \left\{ \Tr \left(-\frac{1}{2} W^{\top} L W \right) \right\} \qquad \mathrm{subject\ to} \qquad W^{\top} W = I
\end{equation}
We propose to solve the constrained optimization problem by means of stochastic gradient optimization.
In view of the orthonormality constraint the elements of $W$ lie on the so called Stiefel manifold, and appealing to theoretical guarantees of convergence of stochastic gradient optimization on manifolds, we can prove that our proposal computes the exact spectrum of $L$ in the limit of the iterations \citep{Bonnabel13}.
In order to simplify the tuning of the optimization procedure, we adapt Adagrad \citep{Duchi11} for stochastic gradient optimization on the Stiefel manifold.
The novelty of our proposal stems from the use of stochastic linear algebra techniques to compute stochastic gradients in $\bigO(n)$.
This leads to computations of stochastic gradient that require processing of a limited number of columns of the full Laplacian matrix, motivating us to name our proposal Mini-Batch Spectral Clustering.

The results on a variety of clustering problems with $n$ up to $580K$ give credence to the value of our proposal.
We can tackle large scale spectral clustering problems achieving the same level of accuracy of the approach that uses the exact spectrum of $L$ at a fraction of the computing time.
We also compare against approximate spectral clustering methods and show that approximations lead to faster solutions that are suboptimal compared to what we can achieve with the proposed method, especially for large data sets\footnote{Code to reproduce all results in the paper is available upon request.}.

\paragraph{Summary of contributions}
(i) We formulate the solution of spectral clustering as a constrained optimization problem that we solve using adaptive stochastic gradient optimization;
(ii) We present a novel way of computing stochastic gradients linearly in the number of data that does not require storing the Laplacian matrix; 
(iii) We analyze the variance of the proposed estimator of the exact gradient to explain the impact of algorithm parameters. 
(iv) We demonstrate that our proposal allows us to tackle large-scale spectral clustering problems by reporting results on data sets of size up to $n = 580K$. Crucially, we can achieve clustering solutions of similar accuracy and orders of magnitude faster compared to the approach that computes the exact spectrum of $L$, and higher accuracy compared to approximate methods at a comparable cost.

\section{Spectral clustering}

\subsection{Background}

Define $X = \{\xvect_1, \ldots, \xvect_n\}$ to be a set of $n$ samples. 
The formulation of spectral clustering introduces an undirected graph $\mathcal{G}$ based on $X$, where the $n$ nodes of $\mathcal{G}$ represent the $n$ input data in $X$, and the edges are weighted according to a similarity measure between the inputs.
The graph $\mathcal{G}$ is expressed by an $n \times n$ adjacency matrix $A$, where each entry $a_{ij}$ determines the weight associated with the edge connecting inputs $i$ and $j$. 
Typically, the elements of the adjacency matrix are defined through off-the-shelf kernel functions, e.g., the Radial Basis Function (RBF) kernel \cite{Scholkopf01}.

Spectral clustering attempts to cluster the elements of $X$ by analyzing the spectral properties of the graph $\mathcal{G}$.
In particular, the objective of spectral clustering is to partition the graph so as to minimize some graph cut criterion, e.g., the normalized cut \citep{Shi00}.
The graph cut problem is generally NP-hard, but its relaxation leads to the definition of the clustering problem as the solution of an algebraic problem \citep{Chung97}.
In particular, following the spectral clustering algorithm proposed by \citet{Ng02}, the graph Laplacian is defined as a normalized version of the adjacency matrix
\begin{equation}
L = D^{-\frac{1}{2}} A D^{-\frac{1}{2}}
\end{equation}
where $D$ is the diagonal matrix of the degrees of the $n$ nodes.
Spectral clustering represents each data point using the corresponding component of the top $k$ eigenvectors of the Laplacian $L$, and computes the solution to the clustering problem by applying $k$-means in this representation.
The difficulty in solving the graph cut problem then becomes calculating the spectrum $L$; this requires $\bigO(n^3)$ computations and $\bigO(n^2)$ space, making it prohibitive -- if not unfeasible -- for large data sets.
The aim of this paper is to address this scaling issue of spectral clustering without sacrificing the accuracy of the solution, as explained next.

\subsection{Spectral clustering as a constrained optimization problem}

The first step to reduce the complexity in finding the top $k$ eigenvectors of $L$, is to cast this algebraic operation as solving the constrained optimization problem in Eq.~\ref{eq:constrained:optimization}.
This is an optimization problem involving $n \times k$ parameters representing the $n$ components of the top $k$ eigenvectors of $L$.
The objective function rewards maximization of a score that, at convergence, is the sum of the $k$ largest eigenvalues.
The constraint $W^{\top} W = I$ gives rise to the so called Stiefel manifold in $\mathcal{R}^{n \times d}$; this imposes orthonormality on the columns of $W$ which, at convergence, represent the eigenvectors associated with the $k$ largest eigenvalues.

The constrained optimization problem in Eq.~\ref{eq:constrained:optimization} can be solved by formulating standard optimization algorithms to deal with the Riemannian metric on the manifold \citep{Smith14}, which typically rely on search operations along geodesics.
Alternative schemes, such as the Cayley transform, have been proposed to tackle optimization problems on Riemannian manifolds \citep{Wen13}.
All these optimization schemes require calculating the gradient of the objective function
\begin{equation} \label{eq:exact:gradient}
G = \nabla_W \Tr \left( -\frac{1}{2} W^{\top} L W \right) = -L W
\end{equation}
This costs $\bigO(n^2)$ and does not require storing $L$ anymore, which is an improvement with respect to computing the full spectrum of $L$.
However, while casting spectral clustering as a constrained optimization problem improves scalability, when $n$ is large it may still take a prohibitive amount of time to be practical.
In the next section we present our proposal to reduce the complexity to solve the constrained optimization problem to $\bigO(n)$ with the guarantee of obtaining the exact solution of the in the limit of the iterations.

\section{Mini-Batch Spectral Clustering (MBSC)}

The intuition behind our proposal is that it is possible to solve the constrained optimization problem in Eq.~\ref{eq:constrained:optimization} relying exclusively on stochastic gradients \cite{Bonnabel13}.
By inspecting the expression of the exact gradient in Eq.~\ref{eq:exact:gradient}, we show how stochastic linear algebra techniques can be employed to unbiasedly estimate exact gradients ($\bigO(n^2)$ cost) with stochastic gradients at $\bigO(n)$ cost.

\subsection{Stochastic optimization on Stiefel manifolds}

 

\begin{figure}[t] 
 \centering
 \includegraphics[trim={2.2cm 2.7cm 0 0}, clip, width=0.28\textwidth]{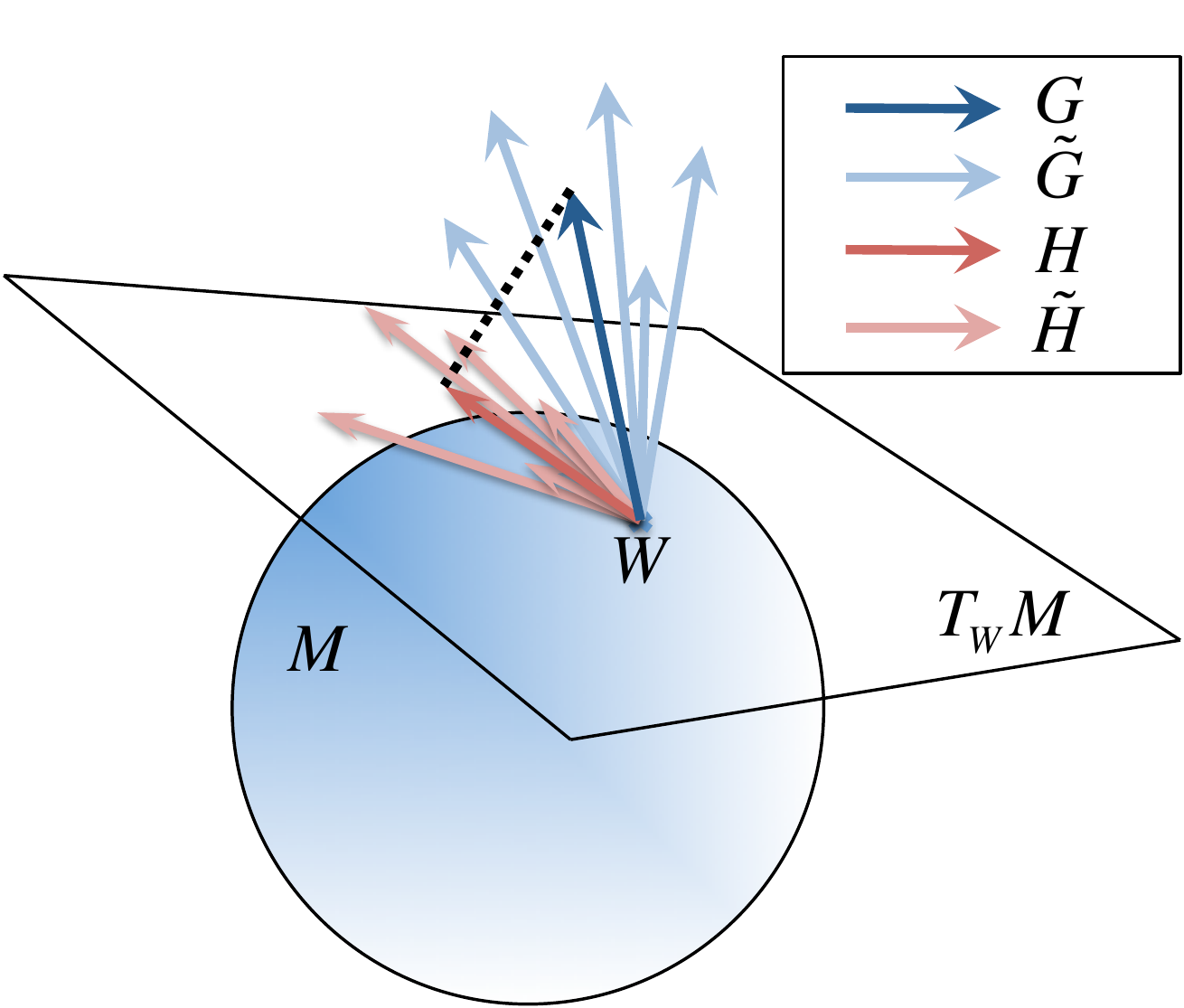} \hspace{2cm}
 \includegraphics[trim={2.2cm 2.7cm 0 0}, clip, width=0.28\textwidth]{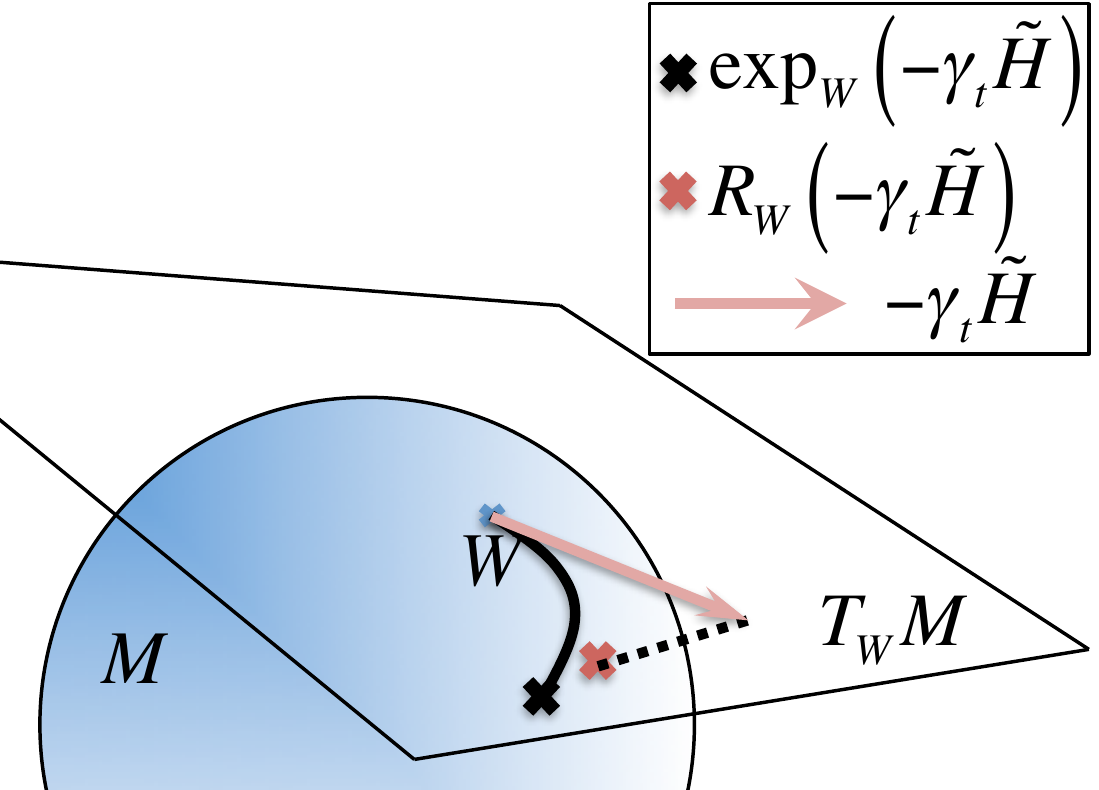}
 \caption{Left: Exact gradient $G$ with a few $\tilde{G}$ and their projection onto the tangent space $T_WM$ of a manifold $M$ at a given $W$ giving $H$ and $\tilde{H}$. Right: Retraction scheme that approximates the exponential map at $W$.}
\label{fig:manifold}
\end{figure}

Stochastic gradient optimization on manifolds is based on the notion of Riemannian gradients, which are elements of the tangent space at a given $W$ that determine the direction of steepest increase of the objective function on the manifold.
For the Stiefel manifold, the Riemannian gradient is \citep{Bonnabel13}
\begin{equation}
H = (I - W W^{\top}) G
\end{equation}
In the case where an unbiased version $\tilde{G}$ of the gradient $G$ is available, namely $\mathrm{E}[\tilde{G}] = G$, an intuitively sensible strategy is to perform stochastic gradient optimization on the manifold.
Formally, this would amount in moving along geodesics for a given length based on the noisy version of the gradient.
Defining $\gamma_t$ as the equivalent of the usual step-sizes in stochastic gradient optimization, and $\exp_{W}()$ as the exponential map at $W$, the update equation would then be $W^{\prime} = \exp_{W}\left(-\gamma_t \tilde{H} \right)$, 
where $\tilde{H} = (I - W W^{\top}) \tilde{G}$ is the unbiased Riemannian gradient of the objective function.
This approach can be shown to converge to the exact solution of the constrained optimization problem in Eq.~\ref{eq:constrained:optimization} \cite{Bonnabel13}.
However, computing the exponential map to simulate the trajectory of the solver on the manifold requires solving potentially expensive differential equations.

An alternative that avoids computing the exponential map altogether is to replace this calculation with an approximation $W^{\prime} = R_{W}\left(-\gamma_t \tilde{H} \right)$ that is much easier to calculate.
If the so called retraction $R_{W}$ satisfies the property that $d(R_W(\delta v), \exp_W(\delta v)) = \bigO(\delta^2)$, where $v$ is an alement of the tangent space, then it is still possible to prove convergence to the exact solution \cite{Bonnabel13}.
A simple and computationally convenient retraction function that satisfies this property is 
\begin{equation}
W^{\prime} = \mathrm{QR}_Q\left(W - \gamma_t (I - W W^{\top}) \tilde{G} \right)
\end{equation}
where $\mathrm{QR}_Q$ extracts the orthonormal factor $Q$ of a $\mathrm{QR}$ decomposition.
This simple retraction moves the optimization in the direction of the stochastic Riemannian gradient $\tilde{H} = (I - W W^{\top}) \tilde{G}$ and applies an orhtonormalization step to ensure that the update is projected back onto the manifold.
Under this choice of retraction, we can appeal to the theoretical results in \citet{Bonnabel13} that ensure convergence to the exact solution in the limit of the iterations similarly to standard stochastic gradient optimization.
An illustration of the retraction scheme is provided in Fig.\ref{fig:manifold}.

\begin{algorithm}[t]
\caption{Stochastic Riemannian gradient on Stiefel Manifold using Mini-Batches} 
\label{alg:Htilde}
\DontPrintSemicolon
\SetKwFunction{Htilde}{Htilde}
\Htilde{$L$, $p$, $N_{\mathbf{r}}$, $W$} { \\ 
Initialize $\tilde{G} \in R^{n\times{k}}$ with elements equal to zero \\
 \For{$i=1$ \KwTo $N_{\mathbf{r}}$}{
   Draw the components of $\mathbf{r}_i$ \\ 
   $\tilde{G}$ += $\frac{1}{N_r}$ $L \mathbf{r}_i \mathbf{r}_i^{\top} W$ 
 }
 Return $\tilde{H}$ = $(I-W W^{\top})\tilde{G}$ 
}
\end{algorithm}

\begin{algorithm}[t]
\caption{Mini-Batch Spectral Clustering}\label{alg:MBSC}
\DontPrintSemicolon
\KwIn{Normalized Laplacian Matrix $L \in R^{n\times{n}}$, number of clusters $k$, regularization factor $\varepsilon$}
\SetKwInOut{Parameter}{Parameters}
\Parameter{Master step length $\lambda$, maximum iteration steps $T$}
\KwOut{Cluster labels of each data points}
Initialize $W^{(0)} \in R^{n\times{k}}$ as a random orthonormal matrix \\ 
Initialize $M^{(0)} \in R^{n\times{k}}$ with elements equal to zero \\
\For{$t=1$ \KwTo $T$}{
      $\tilde{H}^{(t)}$ = \texttt{Htilde}($L$, $p$, $N_{\mathbf{r}}$, $W^{(t-1)}$) \\
      $M_{ij}^{(t)}$ = $M_{ij}^{(t-1)}$ + $|\tilde{H}_{ij}^{(t)}|^{2}$ \\
      $\displaystyle \hat{H}_{ij}^{(t)} = \frac{\tilde{H}_{ij}^{(t)}}{\varepsilon + \sqrt{M_{ij}^{(t)}}}$ \\ 
      $W^{(t)} = W^{(t-1)} - \lambda \hat{H}^{(t)}$ \\
      $W^{(t)} = \mathrm{QR}_Q(W^{(t)})$ \\ 
    }
Apply $k$-means on $W^{(T)}$ to get the cluster labels\\
\end{algorithm}

\subsection{Calculation of stochastic gradients in $\bigO(n)$}

The introduction of stochasticity in the calculation of $\tilde{G}$ follows on from ideas that have been proposed to calculate unbiased stochastic approximations to algebraic quantities, such as traces and log-determinants \citep{Chen11,FilipponeICML15}.
In particular, we define a vector $\mathbf{r}$ such that $\mathrm{E}[\mathbf{r} \mathbf{r}^{\top}] = I$ and we rewrite the expensive matrix product as
\begin{equation}
G = L W = L I W = L \mathrm{E}[\mathbf{r} \mathbf{r}^{\top}] W = \mathrm{E}[ L \mathbf{r} \mathbf{r}^{\top} W ]  
\end{equation}
which suggests that we can replace the exact calculation of $L W$ with the estimator
\begin{equation} \label{eq:stochastic:gradient}
\tilde{G} = \frac{1}{N_r} \sum_{i=1}^{N_r} L \mathbf{r}_i \mathbf{r}_i^{\top} W
\end{equation}

The key to making computations linear in $n$ is to define the components of the random vectors $\mathbf{r}_i$ as drawn from the set $\{-p^{-\frac{1}{2}}, 0, +p^{-\frac{1}{2}}\}$ with probabilities $(p/2, 1 - p, p/2)$ respectively.
It is straightforward to verify that $\mathrm{E}[\mathbf{r}_i \mathbf{r}_i^{\top}] = I$, and $p$ can be chosen to enforce any proportion of zeros in the $\mathbf{r}_i$ vectors.
With this mechanism to inject stochasticity in the calculation of the gradients, we are effectively ignoring some columns of the matrix $L$ whenever there is a zero in the corresponding positions of the $\mathbf{r}_i$ vectors.
This makes it possible to update the parameters $W$ during the solution of the constrained optimization problem in Eq.~\ref{eq:constrained:optimization}, by only selecting a few columns of the full Laplacian.
If the average number of non-zero elements is chosen to be independent of $n$, the calculation of the stochastic gradient is $\bigO(n)$, making the proposed iterative solver linear in the number of samples. 
The memory footprint of the algorithm is a distinctive feature of our proposal; if the degree matrix $D$ is precomputed, calculating the columns of $L$ requires to compute stochastic gradients requires evaluating and normalizing only $\bigO(n)$ elements of the adjacency matrix $A$.

Given that only a subset of the columns in $L$ are used at each iteration to calculate stochastic gradients, we term our proposal Mini-Batch Spectral Clustering (MBSC). 
Instead of defining a probability $p$ to select columns and to repeat this $N_{\mathbf{r}}$ times, we can fix the number and indices of columns that are selected at each iteration (size of the mini-batch) to be $m = l N_{\mathbf{r}}$ and interpret $p = l/n$.
While this is intuitively sensible, fixing the indices of the mini-batches would violate the property that $\mathrm{E}[\mathbf{r}_i \mathbf{r}_i^{\top}] = I$.
One easy way around this issue is to constantly change the way data are split into mini-batches, e.g., by shuffling the data, and this would recover the property $\mathrm{E}[\mathbf{r}_i \mathbf{r}_i^{\top}] = I$.
Even though it is not the focus of the current paper, we envisage the possibility to develop a distributed version of the proposed MBSC algorithm based on our formulation.
The proposed MBSC algorithm is sketched in Algorithms~\ref{alg:Htilde} and~\ref{alg:MBSC}, where,
for the sake of clarity, $L$ is assumed to be stored.
For memory constrained systems where storing the whole Laplacian matrix is unfeasible, our proposal can easily be adapted to avoid storing it, and we report on the performance of this variant in the experiments.


\subsection{Variance of stochastic gradients}

Here we are interested in quantifying the impact of the choice of $p$ and $N_{\mathbf{r}}$ in the calculation of stochastic gradients; to this end, we analyze the variance of the proposed estimator of the exact gradients. 
Without loss of generality, we focus on the variance of a given column of the stochastic gradient $\tilde{G}$, namely the one associated with a given eigenvector, say $\wvect := W_{\cdot s}$.
Recall that the exact gradient with respect to $\wvect$ would be $G_{\cdot s} = L \wvect$ and assume that we use a single vector $\mathbf{r}$ to unbiasedly estimate this as $\tilde{G}_{\cdot s} = L \mathbf{r} \mathbf{r}^{\top} \wvect$.
The covariance of $\tilde{G}_{\cdot s}$ is 
$$
\mathrm{cov}\left(L \mathbf{r} \mathbf{r}^{\top} \wvect \right) = 
\mathrm{E}[ (L \mathbf{r} \mathbf{r}^{\top} \wvect) (L \mathbf{r} \mathbf{r}^{\top} \wvect)^{\top} ] 
- \left(\mathrm{E}[ L \mathbf{r} \mathbf{r}^{\top} \wvect ] \right) \left(\mathrm{E}[ L \mathbf{r} \mathbf{r}^{\top} \wvect ] \right)^{\top}
$$
After some manipulations, that we leave to the supplementary material, we obtain
$$
\diag\left[\mathrm{cov}\left(L \mathbf{r} \mathbf{r}^{\top} \wvect \right)\right] = 
\diag\left[G_{\cdot s} G_{\cdot s}^{\top} \right] + \left(\frac{1}{p} - 3\right) \diag\left[ L \diag(\diag(\wvect \wvect^{\top})) L^{\top} \right] + \diag\left[ L L^{\top} \right]
$$
Given that $\diag(\diag(\wvect \wvect^{\top})) < I$, then $\diag\left[ L \diag(\diag(\wvect \wvect^{\top})) L^{\top} \right] \leq \diag\left[ L L^{\top} \right]$ when $L$ has positive elements.
Also, since $L L^{\top}$ has positive diagonal entries, we can further bound 
\begin{equation}
\diag\left[\mathrm{cov}\left(L \mathbf{r} \mathbf{r}^{\top} \wvect \right)\right] \leq
\diag\left[G_{\cdot s} G_{\cdot s}^{\top} \right] + \frac{1}{p} \diag\left[ L L^{\top} \right]
\end{equation}
The first term contains the square of the components of the exact gradient that will vanish at convergence.
The second term depends on the choice of $p$.
We can rewrite the bound in terms of the mini-batch size $m = l N_{\mathbf{r}}$, for which $p=l/n$, and when $N_{\mathbf{r}}$ vectors $\mathbf{r}_i$ are used to calculate stochastic gradients as in Eq.~\ref{eq:stochastic:gradient}.
\begin{equation}
\diag\left[\mathrm{cov}\left(L \mathbf{r} \mathbf{r}^{\top} \wvect \right)\right] \leq
\frac{1}{N_{\mathbf{r}}}\diag\left[G_{\cdot s} G_{\cdot s}^{\top} \right] + \frac{n}{l N_{\mathbf{r}}} \diag\left[ L L^{\top} \right]
\end{equation}
This reveals that we have two ways of reducing the variance of stochastic gradients; one is to increase $N_{\mathbf{r}}$ and another is to increase $l$.
Imagine that we fix the mini-batch size $m = l N_{\mathbf{r}}$; is it better to increase $l$ and reduce $N_{\mathbf{r}}$, or the other way around?
For the second term it does not matter.
For the first instead, given that it depends only on $N_{\mathbf{r}}$, it is clear that we should favor increasing $N_{\mathbf{r}}$ and reducing $l$.
This entails that we should consider averaging stochastic gradients over several subsets of a mini-batch instead of a few large ones.
This result is interesting because in other popular mini-batch approaches increasing the mini-batch size or the number of repetitions is equivalent.
In our proposed MBSC, because of the nonlinearity of the estimator with respect to the vectors $\mathbf{r}_i$, the bound on the variance shows an unintuitive asymmetry between $N_{\mathbf{r}}$ and $l$. 
A further consideration we can make is that this suggestion is most useful during the first phase of the optimization; towards convergence, the first term will be small and dominated by the second term that is inversely proportional to the mini-batch size $m$.

\section{Experiments}

Throughout the experiments, we make use of the Radial Basis Function (RBF) adjacency function:
$$
a_{ij} = \exp\left( -\frac{\| \xvect_i - \xvect_j \|}{\sigma^2} \right) \qquad i \neq j
$$
where we set $a_{ij} = 0$ when $i = j$.
The RBF adjacency function assigns a large weight to the edge connection inputs that are close in the input space, whereas it assigns a small weight to edges connecting inputs that are far apart.
The parameter $\sigma$ determines the rate of decay of the adjacency function, which can be tuned, e.g., using local statistics on the distances between pairs of points \cite{Zelnik04selftuning}.


We assess clustering performance using the normalized mutual information (NMI) score between the cluster labels and the ground truth class labels.
To reliably measure computational cost of all methods involved in the comparison, we count the amount of floating-point addition and multiplication operations they require, given the affinity matrix, and we also report running time statistics. 
We implemented all the algorithms in Python using the $\texttt{numpy}$ and $\texttt{scikit-learn}$ packages. 
All our experiments are conducted under Ubuntu Linux $14.04$ with $10$-core CPU and $20\mathrm{GB}$ memory. 

Table~\ref{tab:database} summaries the statistics of the data sets, taken from LibSVM \citep{libSVM}, that we consider in the experiments. 
To construct the \texttt{Covtype-I} set, we randomly sample $14129$ samples from the first two classes of the original \texttt{Covtype} data set and merge the data samples of classes $4$, $5$ and $6$ into one single class. 
The purpose is to avoid severe imbalances between classes. 
We make use of all data samples of the original \texttt{Covtype} data set to build the \texttt{Covtype-II} set.  

We organize the experimental study in two parts. 
In both parts, we aim to show the performance of the proposed MBSC algorithm against other state-of-the-art algorithms for solving spectral clustering.
In particular, we compare our proposal with the solution of the power iteration-based algorithm in \citet{Boutsidis15} and the Nystr\"om approximation-based spectral clustering in \citet{Fowlkes04}.
In the first part, this comparison is carried out on moderately large data sets comprising $10K$, $58K$ and $60K$ samples.
For these, we can also report the performance of the spectral clustering approach in \citet{Shi00} where the spectrum of the normalized Laplacian is computed exactly (denoted as ``Exact'' in the plots).
In the second part, we repeat the same comparison on two larger sized data sets, comprising $100K$ and $580K$ samples, where we cannot compute the exact spectrum of $L$.

\begin{table}[t]
\centering
\caption{Summary of the characteristics of the data sets considered in the experiments.}
\label{tab:database}
\begin{tabular}{|c|c|c|c|c|}
\hline
Data set & $\#$ of samples & $\#$ of features & $\#$ of classes & $\sigma$ \\ \hline
 \texttt{Pendigits} & 10992 & 16 & 10 & 223.61\\ \hline
 \texttt{Shuttle} & 58000 & 9 & 7 & 0.45\\ \hline
 \texttt{MNIST} & 60000 & 780 & 10 & 4.08 \\ \hline
 \texttt{Covtype-I} & 100000 & 54 & 5 & 1.15\\ \hline
 \texttt{Covtype-II} & 581012 & 54 & 7 & 1.15\\ \hline
\end{tabular}
\end{table}




\subsection{Comparative evaluation}
We conduct a comparative evaluation of the proposed MBSC with state-of-the-art spectral clustering algorithms on the \texttt{Pendigits}, \texttt{Shuttle} and \texttt{MNIST} data sets \citep{libSVM}. 
In the proposed MBSC approach, we experiment with different choices of the mini-batch size to assess its impact on performance. 
To comprehensively analyze the performance of the proposed method, the iterative stochastic gradient descent on the manifold runs until we make one full pass through the whole data set. 
In practice, the stochastic gradient descent process can be stopped either if the difference between the spectral embedding $W$ derived at two successive iteration steps is less than a threshold, or if a fixed number of iteration is achieved.
For the Nystr\"om approximation, we report a few choices on the number of samples selected to construct the approximate eigenvectors, namely $10$, $50$, $100$, $500$ and $1000$.

\begin{figure}[t]
  \centering
    \includegraphics[width=0.32\textwidth,height=0.32\textwidth]{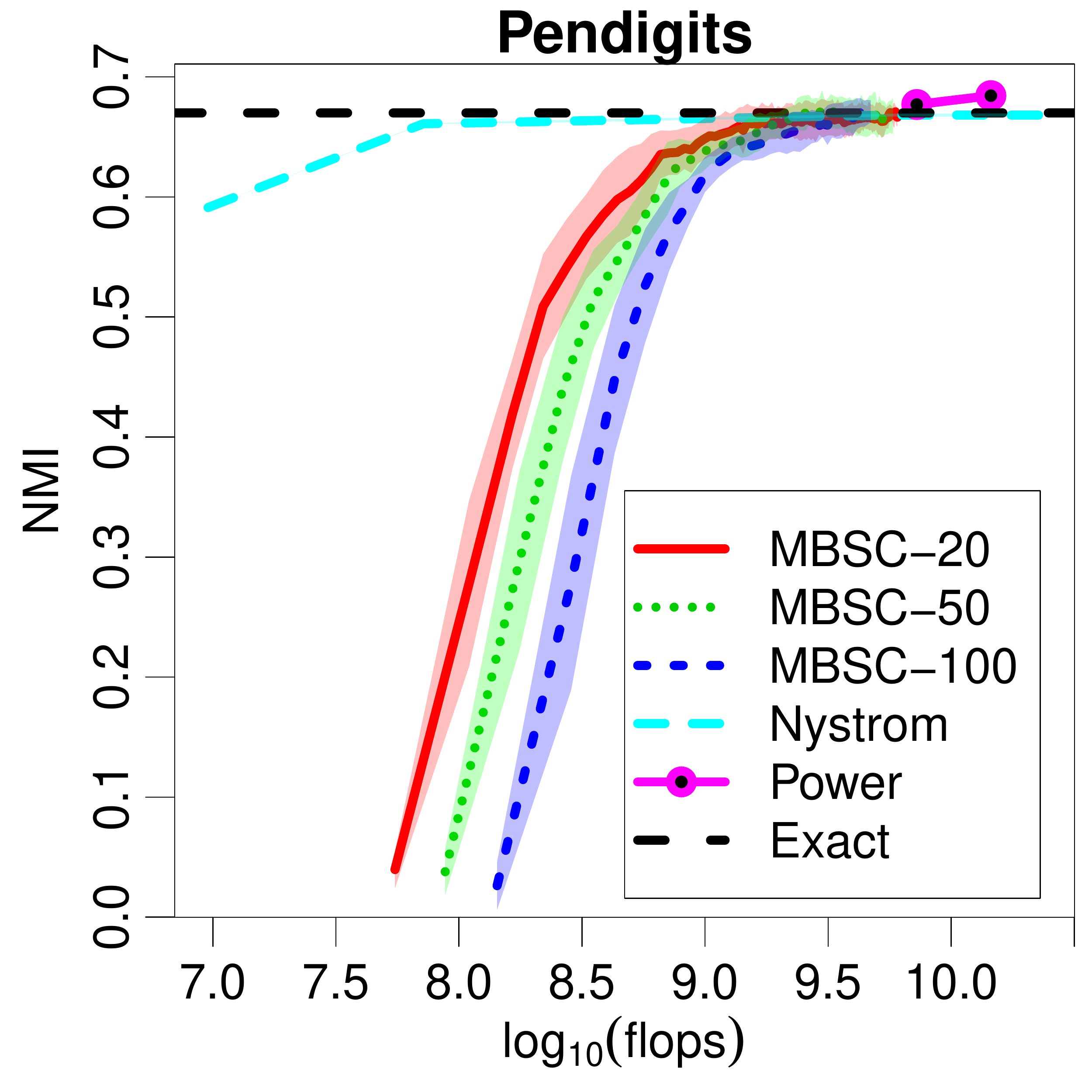}
    \includegraphics[width=0.32\textwidth,height=0.32\textwidth]{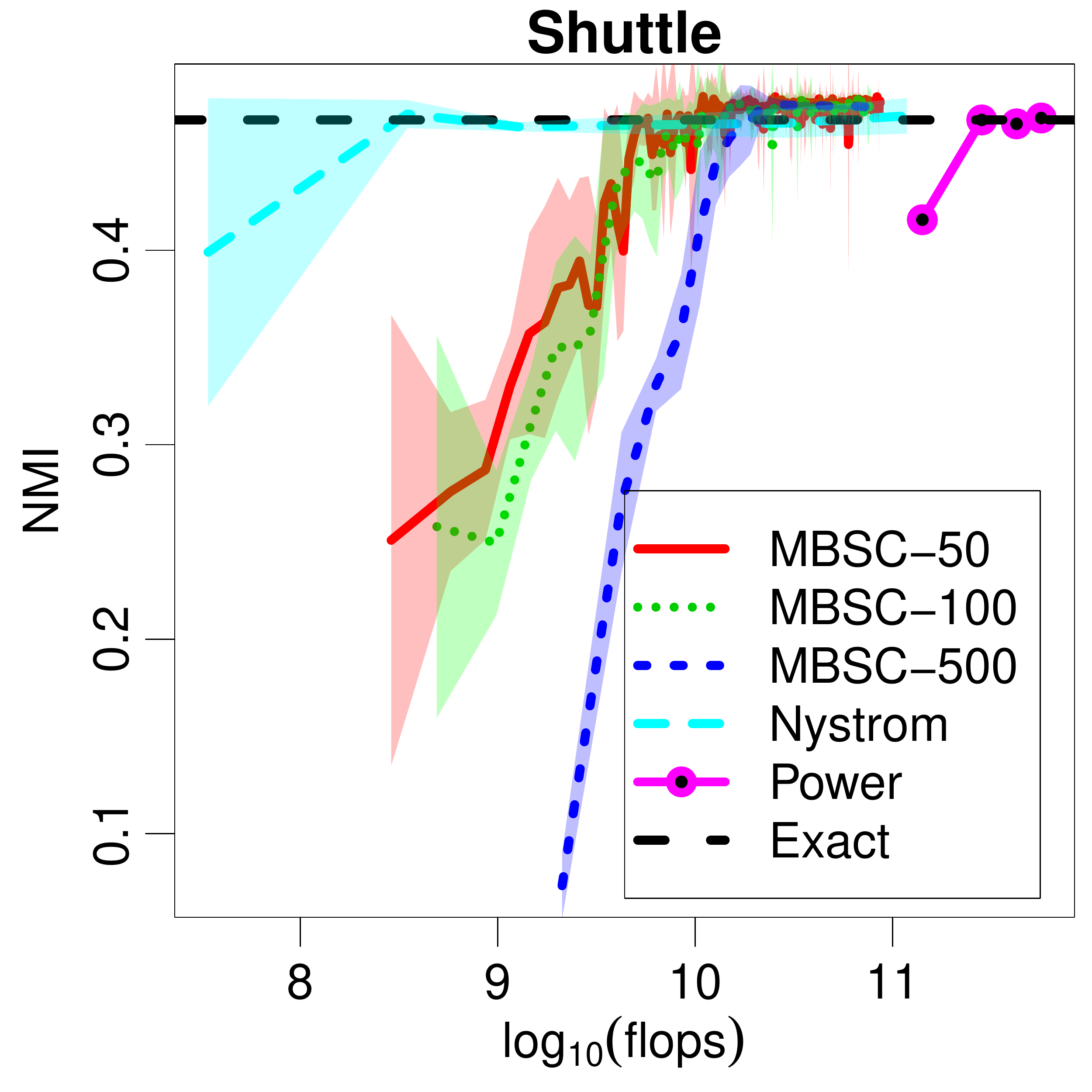}
    \includegraphics[width=0.32\textwidth,height=0.32\textwidth]{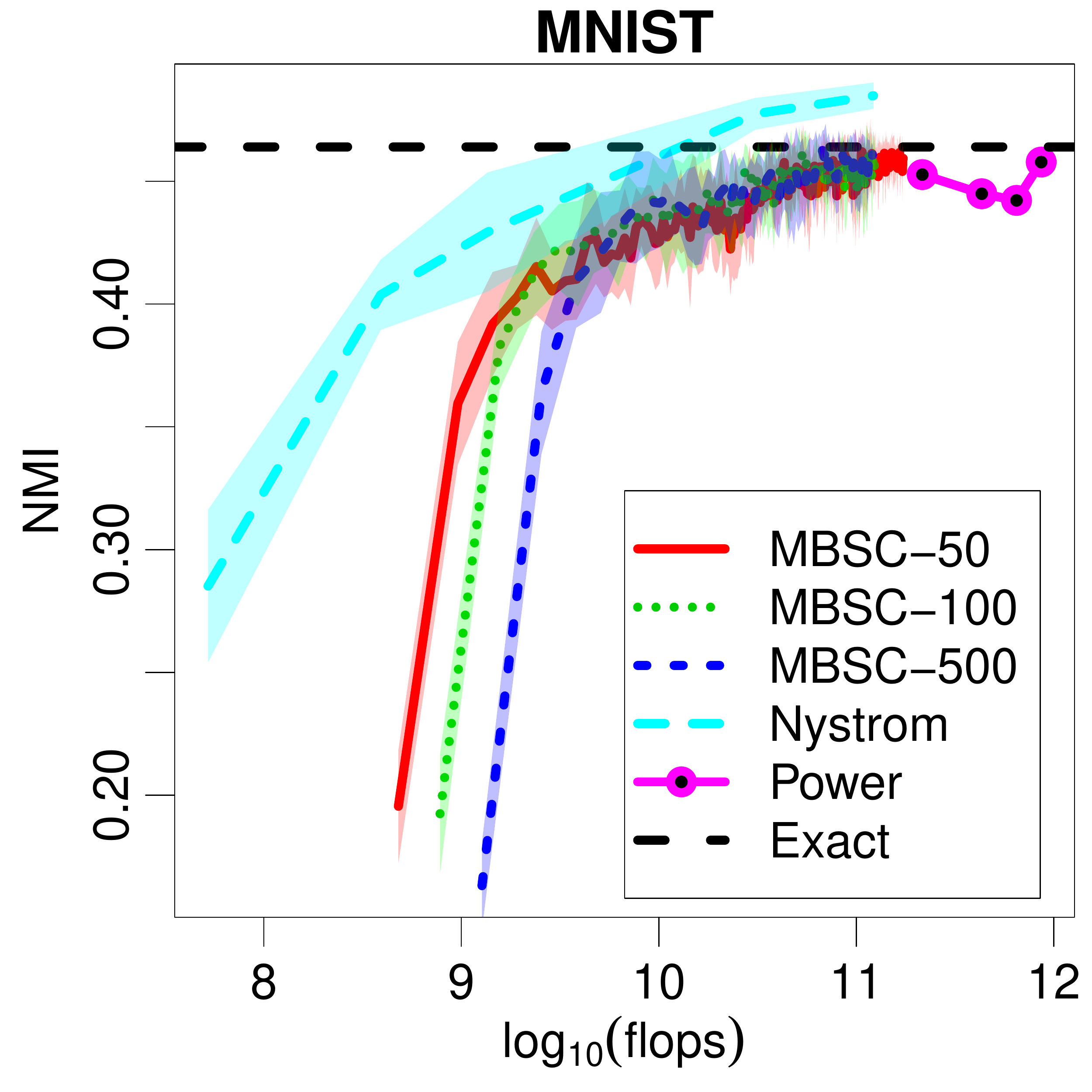}
  \caption{
NMI versus counts of floating-point operations on the \texttt{Pendigits}, \texttt{Shuttle} and \texttt{MNIST} data sets.
NMI for the exact spectral clustering is shown as a constant dashed line. 
MBSC with the largest mini-batches runs for $3.8s$, $107s$ and $112s$ to achieve an average NMI score of $0.67$, $0.48$ and $0.47$ on the three data sets, whereas the power method takes $20s$, $935s$ and $987s$ to get a stable clustering output. 
The Nystr\"om approximation needs $1000$ samples to obtain the optimal clustering accuracy, which takes $15.2s$, $198s$ and $211s$ on the three data sets. 
}\label{fig:comparison}
\end{figure}

Figure~\ref{fig:comparison} illustrates the NMI scores of the proposed MBSC algorithm, the power method, and the Nystr\"om approximation versus the amount of floating-point operations.
In the figure, for the proposed MSBC and the Nystr\"om approximation we report the average plus and minus one standard deviation of the NMI score over $10$ repetitions. 

Recalling that $m$ is the mini-batch size, $k$ is the number of top eigenvectors and the number of clusters, each iteration of MBSC 
requires $2nkm + 6nk^2 - k^2$ floating point operations. 
The power method starts by generating an $n$-by-$k$ Gaussian random matrix $S$, which costs $\bigO(nk)$ operations.
It then computes a matrix product between the $n$-by-$n$ affinity matrix and $S$, and iteratively applies the same multiplication for a total cost of 
$4{n^2}k - 2nk$ floating point operations. 
The final step of the power method performs Singular Value Decomposition (SVD) on an $n$-by-$k$ matrix. 
Since $n{\gg}k$, we adopt the estimate of SVD complexity in \citep{Golub96}, which costs $2nk^2 + 2k^3$ operations. 
Finally, in order to calculate the number of floating point operations for the Nystr\"om approximation we follow the pseudo code in \citep{Fowlkes04},
for which the total count of floating point operations is $6nk+8k^3-3k^2+4n{k^2}-3k+2nkm+nm+2n{m^2}+m^2+m^3-n$.


In Figure~\ref{fig:comparison}, we can observe how the variance of the clustering performance of MSBC diminishes throughout iterations. 
Larger mini-batches lead to faster variance reduction of the stochastic gradient, thus producing faster convergence to the solution.
Compared with the power method, the proposed MBSC needs distinctively less computations, while achieving higher or similar clustering accuracy. 
Another interesting observation is that the proposed MBSC achieves stable clustering accuracy before it makes a full pass through the whole data set. 
The Nystr\"om approximation is computationally fast on all three data sets. 
Nevertheless, its time and space complexity increase drastically if the number of inducing points increases. 
On the \texttt{Pendigits} and \texttt{Shuttle} data sets, with less running time, the proposed MBSC method requires smaller mini batches to conduct clustering and achieves better clustering performance than the Nystr\"om approximation. 
On the \texttt{MNIST} data set, instead, the Nystr\"om approximation produces strikingly good clustering results when the number of selected samples is larger than $500$.
However, the proposed MBSC method converges to the clustering accuracy of the exact spectral clustering even when the size of the mini-batch size is small, e.g., less than $100$. 

\subsection{Use case: spectral clustering on larger data sets}

\begin{table}[t]
\centering
\caption{Running time comparison on \text{Covtype-I} and \text{Covtype-II} data. $T$ is the number of iterations.}
\label{tab:covtype}
\begin{tabular}{|c|c|c|c||c|c|c|c|}
\hline
\multicolumn{4}{|c||}{\texttt{Covtype-I}} & \multicolumn{4}{|c|}{\texttt{Covtype-II}} \\
\hline
Algorithm & $T$ & NMI & time (s) & Algorithm & $T$ & NMI & time (s) \\ \hline
 MBSC-E-400 & 200 & 0.40 & 998  & MBSC-E-1000 & 200 & 0.14 & 7500  \\ \hline
 MBSC-E-800 & 200 & 0.40 & 2100 & MBSC-E-2000 & 200 & 0.14 & 12610 \\ \hline
 Nystrom-400 & - & 0.38 & 78   & Nystrom-1000 & - & 0.09 & 2520   \\ \hline
 Nystrom-800 & - & 0.38 & 965  & Nystrom-2000 & - & 0.11 & 11420   \\ \hline
 Power method & 3 & 0.40 & 9300 & Power Method & \multicolumn{3}{|c|}{Too expensive}  \\ \hline
\end{tabular}
\end{table}

To demostrate that the proposed approach can tackle large-scale spectral clustering problems, we implemented the proposed MBSC algorithm without storing the Laplacian matrix, and applied it on two data sets comprising $100K$ and $580K$ samples. 
This variant of the code, that we denote as MBSC-E, computes the necessary columns of $L$ to construct stochastic gradients on-the-fly.


Table~\ref{tab:covtype} reports the overall running time and NMI of MBSC-E on the two data sets, where MBSC-E produces stable clustering results after $200$ iterations. 
In addition, we compare against the Nystr\"om approximation and the power method. 
In the comparison, the Nystr\"om approximation selects a subset of the same size of the mini-batch in MBSC-E. 
On the \texttt{Covtype-II} data set, the power method fails to obtain clustering results within an acceptable time, and we omit it from Table~\ref{tab:covtype}. 

While running time is heavily dependent on implementation and system architecture, we argue that this is probably in favor of the Nystr\"om approximation, for which we are using well optimized scientific computing packages.
In any case, the purpose of this experiment is to demonstrate that the proposed MBSC algorithm, being exact in the limit of iterations, can achieve higher performance than approximate methods on large scale spectral clustering problems.
Crucially, we demonstrate that this is possible at a comparable computational cost with approximate methods.

On both \texttt{Covtype-I} and \texttt{Covtype-II} data sets, MBSC achieves consistently better clustering accuracy than the Nystr\"om approximation. 
Because the computational cost of the Nystr\"om approximation rapidly increases with the size of the approximating set, it requires longer than MBSC-E to achieve a comparable clustering accuracy. 
Furthermore, compared with the power method, MBSC-E shows superior computational efficiency for large-scale spectral clustering problems. 
Remarkably, MBSC-E requires less than $1\mathrm{GB}$ and $3.6\mathrm{GB}$ memory to run on the two data sets, respectively. 




\section{Conclusions}
With the aim of improving scalability of spectral clustering, in this work we formulated normalized cut spectral clustering as an optimization problem with an orthonormality constraint that we could solve using stochastic gradient optimization.
We proposed a novel adaptive stochastic gradient optimization framework on Stiefel manifolds to compute the spectrum of Laplacian matrices, with computation of stochastic gradients linear in the number of samples. 
We provided theoretical justifications and empirical analyses to demonstrate how our proposal can tackle large-scale spectral clustering problems in a practical way.

The proposed stochastic optimization is characterized by attractive robustness to parameter selection and scalability properties, leading to the same clustering accuracy of spectral clustering approaches that use the exact spectrum of the Laplacian at a fraction of the cost. 
The results also support the motivation behind our proposal that approximate methods can potentially affect clustering performance.
In cases where approximate methods perform well, as we reported in one of the experiments, we can see our proposal as a practical way to obtain the gold-standard of the ``exact'' approach at a reasonable cost. 
Furthermore, the proposed method does not need to load the whole Laplacian matrix into memory, making it especially suitable for handling large-scale spectral clustering with a limited memory footprint. 


There are a number of extensions that we are currently investigating, such as the possibility to combine our framework with other approximate methods, for example to be able to afford more inducing points when performing spectral clustering using the Nystr\"om approximation.
Another extension is to leverage approximations to reduce the variance of the stochastic gradients without introducing any bias to accelerate stochastic gradient optimization. 
A Spark/TensorFlow implementation of the proposed algorithm is under development.

\subsubsection*{Acknowledgments}

The Authors would like to thank Yun Shen from Symantec Research Labs for his support in setting up the experiments and optimizing the implementation of the algorithms.


\small
 
\begin{thebibliography}{19}
\providecommand{\natexlab}[1]{#1}
\providecommand{\url}[1]{\texttt{#1}}
\expandafter\ifx\csname urlstyle\endcsname\relax
  \providecommand{\doi}[1]{doi: #1}\else
  \providecommand{\doi}{doi: \begingroup \urlstyle{rm}\Url}\fi

\bibitem[Shi and Malik(2000)]{Shi00}
J.~Shi and J.~Malik.
\newblock {Normalized Cuts and Image Segmentation}.
\newblock \emph{{IEEE} Transactions on Pattern Analysis and Machine
  Intelligence}, 22\penalty0 (8):\penalty0 888--905, 2000.

\bibitem[Ng et~al.(2002)Ng, Jordan, and Weiss]{Ng02}
A.~Y. Ng, M.~I. Jordan, and Y.~Weiss.
\newblock {On Spectral Clustering: Analysis and an algorithm}.
\newblock In \emph{Advances in Neural Information Processing Systems 14},
  Cambridge, MA, 2002.

\bibitem[Dhillon et~al.(2004)Dhillon, Guan, and Kulis]{Dhillon04}
I.~S. Dhillon, Y.~Guan, and B.~Kulis.
\newblock {Kernel k-means: spectral clustering and normalized cuts}.
\newblock In \emph{Proceedings of the tenth ACM SIGKDD international conference
  on Knowledge discovery and data mining}, pages 551--556, New York, NY, USA,
  2004.

\bibitem[von Luxburg(2007)]{vonLuxburg07}
U.~von Luxburg.
\newblock {A tutorial on spectral clustering}.
\newblock \emph{Statistics and Computing}, 17\penalty0 (4):\penalty0 395--416,
  2007.

\bibitem[Boutsidis et~al.(2015)Boutsidis, Kambadur, and Gittens]{Boutsidis15}
C.~Boutsidis, P.~Kambadur, and A.~Gittens.
\newblock {Spectral Clustering via the Power Method — Provably}.
\newblock In \emph{{Proceedings of the 32nd International Conference on Machine
  Learning}}, pages 40--48, 2015.

\bibitem[Fowlkes et~al.(2004)Fowlkes, Belongie, Chung, and Malik]{Fowlkes04}
C.~Fowlkes, S.~Belongie, F.~Chung, and J.~Malik.
\newblock {Spectral Grouping Using the Nystr\"{o}m Method}.
\newblock \emph{IEEE Transactions on Pattern Analysis and Machince
  Intelligence}, 26\penalty0 (2):\penalty0 214--225, 2004.

\bibitem[Yan et~al.(2009)Yan, Huang, and Jordan]{Yan09}
D.~Yan, L.~Huang, and M.~I. Jordan.
\newblock {Fast Approximate Spectral Clustering}.
\newblock In \emph{Proceedings of the 15th ACM SIGKDD International Conference
  on Knowledge Discovery and Data Mining}, pages 907--916, New York, NY, USA,
  2009.

\bibitem[Li et~al.(2016)Li, Huang, and Liu]{Li16}
Y.~Li, J.~Huang, and W.~Liu.
\newblock {Scalable Sequential Spectral Clustering}.
\newblock In \emph{Proceedings of the Thirtieth {AAAI} Conference on Artificial
  Intelligence, February 12-17, 2016, Phoenix, Arizona, {USA.}}, pages
  1809--1815, 2016.

\bibitem[Bonnabel(2013)]{Bonnabel13}
S.~Bonnabel.
\newblock Stochastic gradient descent on {R}iemannian manifolds.
\newblock \emph{IEEE Transactions on Automatic Control}, 58\penalty0
  (9):\penalty0 2217--2229, 2013.

\bibitem[Duchi et~al.(2011)Duchi, Hazan, and Singer]{Duchi11}
J.~Duchi, E.~Hazan, and Y.~Singer.
\newblock {Adaptive Subgradient Methods for Online Learning and Stochastic
  Optimization}.
\newblock \emph{Journal of Machine Learning Research}, 12:\penalty0 2121--2159,
  2011.

\bibitem[Sch\"{o}lkopf and Smola(2001)]{Scholkopf01}
B.~Sch\"{o}lkopf and A.~J. Smola.
\newblock \emph{{Learning with Kernels: Support Vector Machines,
  Regularization, Optimization, and Beyond}}.
\newblock MIT Press, Cambridge, MA, USA, 2001.

\bibitem[Chung(1997)]{Chung97}
F.~R.~K. Chung.
\newblock \emph{{Spectral Graph Theory (CBMS Regional Conference Series in
  Mathematics, No. 92)}}.
\newblock American Mathematical Society, 1997.

\bibitem[Smith(2014)]{Smith14}
S.~T. Smith.
\newblock {Optimization Techniques on {R}iemannian Manifolds}, 2014.
\newblock arXiv:1407.5965.

\bibitem[Wen and Yin(2013)]{Wen13}
Z.~Wen and W.~Yin.
\newblock {A feasible method for optimization with orthogonality constraints}.
\newblock \emph{Mathematical Programming}, 142\penalty0 (1):\penalty0 397--434,
  2013.

\bibitem[Chen et~al.(2011)Chen, Anitescu, and Saad]{Chen11}
J.~Chen, M.~Anitescu, and Y.~Saad.
\newblock {Computing f(A)b via Least Squares Polynomial Approximations}.
\newblock \emph{SIAM Journal on Scientific Computing}, 33\penalty0
  (1):\penalty0 195--222, 2011.

\bibitem[Filippone and Engler(2015)]{FilipponeICML15}
M~Filippone and R.~Engler.
\newblock Enabling scalable stochastic gradient-based inference for {G}aussian
  processes by employing the {Unbiased} {LInear} {System} {SolvEr} {(ULISSE)}.
\newblock In \emph{{Proceedings of the 32nd International Conference on Machine
  Learning}}, pages 1015--1024, 2015.

\bibitem[Zelnik-Manor and Perona(2004)]{Zelnik04selftuning}
L.~Zelnik-Manor and P.~Perona.
\newblock Self-tuning spectral clustering.
\newblock In \emph{Advances in Neural Information Processing Systems 17}, pages
  1601--1608, 2004.

\bibitem[Chang and Lin(2011)]{libSVM}
C.-C. Chang and C.-J. Lin.
\newblock {LIBSVM}: A library for support vector machines.
\newblock \emph{ACM Transactions on Intelligent Systems and Technology},
  2:\penalty0 27:1--27:27, 2011.

\bibitem[Golub and Van~Loan(1996)]{Golub96}
G.~H. Golub and C.~F. Van~Loan.
\newblock \emph{{Matrix computations}}.
\newblock The Johns Hopkins University Press, 3rd edition, 1996.

\end{thebibliography}


\newpage

\normalsize

\appendix

\section{Derivation of the variance of stochastic gradients}

Recall that the covariance of the stochastic gradient is:
$$
\mathrm{cov}\left(L \mathbf{r} \mathbf{r}^{\top} \wvect \right) = 
\mathrm{E}[ (L \mathbf{r} \mathbf{r}^{\top} \wvect) (L \mathbf{r} \mathbf{r}^{\top} \wvect)^{\top} ] 
- \left(\mathrm{E}[ L \mathbf{r} \mathbf{r}^{\top} \wvect ] \right) \left(\mathrm{E}[ L \mathbf{r} \mathbf{r}^{\top} \wvect ] \right)^{\top}
$$
By expanding the first term and realizing that the second term in the right hand side is just the outer product of the $s$th column of $\tilde{G}$ with itself, we obtain that the variance of the components of the stochastic gradient is:
$$
\diag\left[\mathrm{cov}\left(L \mathbf{r} \mathbf{r}^{\top} \wvect \right)\right] = 
\diag\left[ L \mathrm{E} \left[ \mathbf{r} \mathbf{r}^{\top} \wvect \wvect^{\top} \mathbf{r} \mathbf{r}^{\top}  \right] L^{\top}   \right]
- \diag\left[L \wvect \wvect^{\top}  L^{\top}\right]
$$

The focus of this supplement is to derive an expression for the expectation
$$
\mathrm{E} \left[ \mathbf{r} \mathbf{r}^{\top} \wvect \wvect^{\top} \mathbf{r} \mathbf{r}^{\top}  \right] 
$$
needed to calculate the variance of the stochastic gradients.
The $ij$ element of the expectation is
\begin{eqnarray}
\mathrm{E}(\mathbf{r} \mathbf{r}^{\top} \wvect \wvect^{\top} \mathbf{r} \mathbf{r}^{\top})_{ij}
& = &
\sum_{kl} \sum_{r_i, r_k, r_l, r_j} r_i r_r w_{k} w_{l} r_s r_j P(r_i) P(r_k) P(r_l) P(r_j) \\
& = & 
\sum_{kl} w_k w_l \sum_{r_i, r_k, r_l, r_j} r_i r_k r_l r_j P(r_i) P(r_k) P(r_l) P(r_j)
\end{eqnarray}
We need to consider all possible $n^4$ cases for the indices going from $1$ to $n$.
\begin{itemize}
\item $i \neq j \neq k \neq l$ - there are $\frac{n!}{(n-4)!}$ of these cases - all variables are independent and the expectation factorizes into the product of the expectations, which is zero.

\item two pairs are equal - there are $3\frac{n!}{(n-2)!}$ of these cases.
It is useful to realize the following expectation:
$$
\sum_{r_i, r_k, r_l, r_j} r_i r_k r_l r_j P(r_i) P(r_k) P(r_l) P(r_j) = 4 \left(\frac{1}{\sqrt{p}}\right)^2 \left(\frac{1}{\sqrt{p}}\right)^2 \frac{p}{2} \frac{p}{2} = 1
$$
Within this set of cases we distinguish three cases, and there are $\frac{n!}{(n-2)!}$ for each of these:
\begin{itemize}
\item $(i = j) \neq (k = l)$ - these are useful for the calculation of the diagonal of the expectation and they give $\sum_{k \neq i} w_k w_{k}$

\item $(i = k) \neq (j = l)$ - these give $w_i w_j$

\item $(i = l) \neq (k = j)$ - these give $w_i w_j$
\end{itemize}

So for the off-diagonal elements of the expectation we have the sum of the last two cases above, giving $2 w_i w_j$.

\item two out of four are equal and the other two are different - there are $\frac{n!}{(n-3)!} \frac{4!}{(4-2)!2!}$ of these cases and they give a zero.

\item three out of four are equal - there are $4 \frac{n!}{(n-2)!}$ of these combinations - these cases give a zero.

\item $i = j = k = l$ - there are $n$ of these combinations - when the indices are all the same, we have nonzero in the two cases of $v$ being $-\frac{1}{\sqrt{p}}$ or $+\frac{1}{\sqrt{p}}$, giving
$$
2 w_i w_i \left(\frac{1}{\sqrt{p}}\right)^4 \frac{p}{2} = \frac{w_i w_i}{p}
$$
\end{itemize}

We are now ready to compute the expectation above.
We distinguish two cases:
\begin{itemize}
\item $i \neq j$ - the off-diagonal of the expectation which is:
$$
2 w_i w_j
$$
\item $i = j$ - the diagonal of the expectation which is:
$$
\frac{w_i w_i}{p} + \sum_{k \neq i} w_{kk}
$$
\end{itemize}

Because of these expressions, the expectation can be rewritten in matrix form as:
$$
\mathrm{E}(\mathbf{r} \mathbf{r}^{\top} \wvect \wvect^{\top} \mathbf{r} \mathbf{r}^{\top}) = 2 \wvect \wvect^{\top} - 2 \diag(\diag(\wvect \wvect^{\top})) + \frac{1}{p} \diag(\diag(\wvect \wvect^{\top})) + \Tr(\wvect \wvect^{\top}) I - \diag(\diag(\wvect \wvect^{\top}))
$$
We realize that
$$
\Tr(\wvect \wvect^{\top}) = \| \wvect \|^2 = 1
$$
because of the orthonormality constraint, so after substituting this expression and gathering terms we obtain
$$
\mathrm{E}(\mathbf{r} \mathbf{r}^{\top} \wvect \wvect^{\top} \mathbf{r} \mathbf{r}^{\top}) = 2 \wvect \wvect^{\top} + \left(\frac{1}{p} - 3\right) \diag(\diag(\wvect \wvect^{\top})) + I
$$
Plugging this into the expression of the variance of the components of the stochastic gradient we obtain 
$$
\diag\left[\mathrm{cov}\left(L \mathbf{r}_i \mathbf{r}_i^{\top} \wvect \right)\right] = 
\diag\left[ L \left( 2 \wvect \wvect^{\top} + \left(\frac{1}{p} - 3\right) \diag(\diag(\wvect \wvect^{\top})) + I \right) L^{\top}   \right]
- \diag\left[L \wvect \wvect^{\top}  L^{\top}\right]
$$
This can be simplifed into
$$
\diag\left[\mathrm{cov}\left(L \mathbf{r}_i \mathbf{r}_i^{\top} \wvect \right)\right] = 
\diag\left[ L \left( \wvect \wvect^{\top} + \left(\frac{1}{p} - 3\right) \diag(\diag(\wvect \wvect^{\top})) + I \right) L^{\top}   \right]
$$
and finally into
$$
\diag\left[\mathrm{cov}\left(L \mathbf{r}_i \mathbf{r}_i^{\top} \wvect \right)\right] = 
\diag\left[ L \wvect \wvect^{\top} L^{\top} + \left(\frac{1}{p} - 3\right) L \diag(\diag(\wvect \wvect^{\top})) L^{\top} + L L^{\top} \right]
$$

\end{document}